\documentclass[10pt,twocolumn,letterpaper]{article}

\usepackage[pagenumbers]{cvpr} 

\usepackage{graphicx}
\usepackage{amsmath}
\usepackage{amssymb}
\usepackage{booktabs}
\usepackage{subfigure}

\usepackage{soul}
\usepackage{multirow}
\usepackage[pagebackref,breaklinks,colorlinks]{hyperref}

\usepackage[capitalize]{cleveref}
\crefname{section}{Sec.}{Secs.}
\Crefname{section}{Section}{Sections}
\Crefname{table}{Table}{Tables}
\crefname{table}{Tab.}{Tabs.}

\def\cvprPaperID{*****} 
\def\confName{CVPR}
\def\confYear{2022}

\usepackage{overpic}
\usepackage{enumitem} 
\usepackage{overpic} 
\usepackage{color}

\definecolor{turquoise}{cmyk}{0.65,0,0.1,0.3}
\definecolor{purple}{rgb}{0.65,0,0.65}
\definecolor{dark_green}{rgb}{0, 0.5, 0}
\definecolor{orange}{rgb}{0.8, 0.6, 0.2}
\definecolor{red}{rgb}{0.8, 0.2, 0.2}
\definecolor{darkred}{rgb}{0.6, 0.1, 0.05}
\definecolor{blueish}{rgb}{0.0, 0.3, .6}
\definecolor{light_gray}{rgb}{0.7, 0.7, .7}
\definecolor{pink}{rgb}{1, 0, 1}
\definecolor{greyblue}{rgb}{0.25, 0.25, 1}

\usepackage{blindtext}

\renewcommand{\paragraph}[1]{\vspace{1em}\noindent\textbf{#1}.}
\begin{document}
\title{Open World Detection for One Stage Detection and its impact on Classification \textcolor{red}{ need a better name}}

\title{Extending One-Stage Detection with Open-World Proposals}

\author{Sachin Konan\\
Georgia Institute of Technology\\
{\tt\small skonan3@gatech.edu}
\and
Kevin J Liang, Li Yin\\
Facebook AI \\
{\tt\small \{kevinjliang, liyin\}@fb.com}

}
\maketitle
\begin{abstract}
In many applications, such as autonomous driving, hand manipulation, or robot navigation, object detection methods must be able to detect objects unseen in the training set. Open World Detection (OWD) seeks to tackle this problem by generalizing detection performance to seen and unseen class categories. Recent works have seen success in the generation of class-agnostic proposals, which we call Open-World Proposals (OWP), but this comes at the cost of a big drop on the classification task when both tasks are considered in the detection model. These works have investigated two-stage Region Proposal Networks (RPN) by taking advantage of objectness scoring cues; however, for its simplicity, run-time, and decoupling of localization and classification, we investigate OWP through the lens of fully convolutional one-stage detection network, such as FCOS~\cite{tian2019fcos}. We show that our architectural  and  sampling optimizations on FCOS can increase OWP performance by as much as $6\%$ in recall on novel classes, marking the first proposal-free one-stage detection network to achieve comparable performance to RPN based two-stage networks. Furthermore, we show that the inherent, decoupled architecture of FCOS has benefits to retaining classification performance. While two-stage methods worsen by 6\% in recall on novel classes, we show that FCOS only drops 2\% when jointly optimizing for OWP and classification.
\end{abstract}
\section{Introduction}
\label{sec:intro}

Object detection consists of two tasks: localization and detection. Many prior works have performed this in two stages\cite{ren2016faster, lin2017feature, he2018mask, vu2019cascade, lin2018focal}: the localization stage finds bounding boxes of potential objects, which are then fed to a classification stage that classifies each object within a finite set of categories. In Faster R-CNN~\cite{ren2016faster}, for example, the classification stage is conditioned on the boxes generated from a region proposal network (RPN) and have achieved great success on several datasets; however, despite their success, these networks have shown limitations in detecting objects with large sizing variability and require exhaustive tuning for cross-dataset generalization \cite{rcnn_survey, zhong2018anchorfree, xin2021pafnet}. Furthermore, by conditioning the classification stage on localization, the proposals generated by the localization stage tend to over-fit the training class categories, so they tend not to be accurate on objects unavailable in the training set \cite{cross_dataset, fan2021generalized}. We seek to solve this issue by addressing Open World Detection (OWD), and specifically consider its class-agnostic proposal generation sub-problem \cite{Uijlings2013, wang2020leads, kim2021learning}, which we refer to as \textit{Open-World Proposals (OWP)}. Generating OWP
would spawn more effective localization networks that can be applied to tasks such as detecting unknown objects during autonomous driving \cite{GUPTA2021100057}, automatic threat detection \cite{liang2019toward}, long-tail vocabulary detection \cite{gupta2019lvis}, or class-agnostic semantic segmentation \cite{zhang2019canet}. We also consider OWP as a promising meta-learning problem because proposal generation is applicable to any specific object detection dataset.

Object detection datasets only provide ground-truth boxes of certain objects in the foreground, and we seek to train on these annotations to learn fundamental characteristics of objects that can generalize well to unseen class categories. Prior works have researched the efficacy of objectness cues such as centerness, intersection over union (IOU)\cite{tian2019fcos, jiang2018acquisition, lin2018focal} to mask logit predictions that are not ``near'' the ground-truth annotations. Object Localization Network (OLN)\cite{kim2021learning} showed that combining these cues with an RPN made it capable of generating class-agnostic proposals. When training a class-agnostic detector, we split the training class categories into two sets: base classes which are used for training and novel unseen classes, which are used during evaluation to test the generalizability to new object types. Following OLN, we seek to maximize the OWP performance on the unseen class category set. However, in contrast to OLN, we also seek to maintain classification performance on the seen class category set, which we call \textit{OWP-aware Classification}. This is particularly important, because while generating OWP are important for many applications, it is often critical to still be able to classify a subclass of relevant categories \cite{cross_dataset}.

Previous works have explored two-stage network~\cite{kim2021learning, joseph2021open, jiang2018acquisition, cat_agnostic_analysis} architectures; however, one-stage detection networks \cite{tian2019fcos, redmon2016look, lin2018focal, ssd} have increasingly been investigated for their comparable performance, faster inference speeds, and simplicity in relation to two-stage networks \cite{lu2020mimicdet}. We chose to investigate fully-convolutional one-stage detection (FCOS)~\cite{tian2019fcos} for its simplicity, but more importantly, its architectural potential to achieve state-of-the-art OWP performance while still maintaining training-class-specific classification accuracy. The FCOS architecture shows promise because it decouples classification from localization, so while two-stage detectors' RPNs will be affected by the gradients of the downstream classifier \cite{gradients, kim2021learning}, we hypothesize that the FCOS regression head won't directly suffer from any classification losses and vice versa\cite{xu2019survey, cabana2021backward}. This provides the potential to maintain classification performance while still making OWPs. Additionally, the regression head already has a centerness\cite{tian2019fcos} objectness prediction head to emphasize proposals closer to the center of the ground-truth proposals; however, this baseline version of FCOS has poor performance on the OWP task. We improve FCOS's OWP ability by investigating IOU \cite{measuringobjectness, lin2018focal} as an objectness cue, as well as different sampling methods that can create more balanced sets of examples to train the objectness branches. Finally, we investigate OWP-aware classification by training fine-tuned models from the OWP task, as well as joint training the classification and regression heads in tandem. We also propose \textit{Unknown Object Masking}, to remove incorrectly labeled background samples~\cite{yang2020object} during classification training.

We test our optimizations on both the COCO and LVIS datasets. Our sampling/objectness optimizations improve baseline OWP performance by \textbf{+4.8\%} and \textbf{+4.55\%} on COCO and LVIS respectively, outperforming one-stage OLN. During the OWP-aware classification, our optimizations only reduce OWP performance by \textbf{-2\%} and \textbf{-0.57\%} on COCO and LVIS, respectively, whereas OLN drops by \textbf{-6.2\%} on COCO.

Our work is the first known application of a one-stage detection network, like FCOS, to the OWP task, whose novelty comes from a more thorough empirical evaluation of various architectural/sampling optimizations to improve OWP and OWP-aware classification performance. Additionally, we investigate the pitfalls of standard FCOS implementations and discuss their effects on both tasks. These contributions can be summarized as follows:
\begin{enumerate}
    \item \textbf{IOU Overlap as Objectness } The default FCOS implementation uses centerness as an objectness cue \cite{tian2019fcos}; however, we investigate using IOU and the combination of IOU and centerness (similar to OLN) for box scoring. Additionally, we study the impact of conditioning the IOU/Centerness branches on both the Feature Pyramid and the regression head for enhancing OWP performance.
    \item \textbf{Sampling Procedures } For training the regression head, FCOS firstly only considers foreground pixels, which are then sampled if they are within some radial distance to the center of the ground-truth and if the size of the bounding box at that pixel is within the bounds of the Feature Pyramid~\cite{lin2017feature}. We analyze the influence of these sampling procedures on training the objectness branches and propose a new masking procedure called \textit{IOU Sampling} to introduce more hard negatives and balance the training set.
    \item \textbf{Unknown Object Masking } For training the classification head, default FCOS considers all foreground pixels and their respective labels, and every other pixel as part of background. In reality, many images are non-exhaustively annotated~\cite{yang2020object}, so to minimize mis-classified background pixels, we propose Unknown Object Masking, which uses the predictions from the objectness branches to mask background pixels that ``look'' like they belong to an object.
\end{enumerate}

\section{Related works}
\label{sec:related}

There has been recent renewed interest in OWD, although there have been many previous works that have investigated the extension of object detection networks to unknown or unseen class categories. Additionally, early computer vision was focused on class-agnostic OWD, or general object proposal generation. Below we outline previous works in proposal generation, OWD, and additionally introduce the details of the FCOS architecture.

\paragraph{Class-Agnostic Proposals}
Early computer vision attempted to solve what is essentially the OWP paradigm, class-agnostic proposal generation, by applying techniques like edge detection, gaussian filters, and more, to discover bounding boxes around objects \cite{measuringobjectness, cat_agnostic_analysis,combo_grouping}. These works attempt to create heuristics that quantitatively capture the generalized characteristics of objects; however, through the evolution of deep learning, learning-based methods have replaced heuristic-driven algorithms due to better performance \cite{girshick2014rich, gidaris2016attend}. There exist two main types of detection networks: one-stage and two-stage architectures.

As previously stated, two-stage networks \cite{ren2016faster, lin2017feature, he2018mask, vu2019cascade, lin2018focal} have seen much success in the proposal generation regime, because of the Region Proposal Network (RPN) that makes proposals that are used by downstream classification tasks. Many supplementary works have been conducted that improve the efficacy of these proposals \cite{vu2019cascade, small_object_2017} by improving their accuracy and reducing redundancy. However, while these methods sport improvements on the training class categories, they don't generalize well to unseen class categories. There has been another set of research that has tried to investigate the long-tail distribution problem \cite{zhong2021improving, ren2020balanced, zhang2021distribution, liu2019largescale}, which attempts to solve generalized object detection with datasets that are heavily left-skewed, i.e. there are a small number of highly frequent categories and a large number of low frequency categories. Liu \etal \cite{liu2019largescale} considered the long-tail distribution problem, and they developed an attentional architecture capable of generating generalizable feature embeddings that had good classification accuracy over both low-frequency/unknown classes. Both Liu \etal and Bendale \etal \cite{liu2019largescale, bendale2015open} considered the open-set problem and supplemented both the logit output and the training/validation sets with ``unknown'' classes. These show good performance, but previous OWD-specific works \cite{kim2021learning, joseph2021open} differ by not supplementing the classifier with an ``unknown label'' or even ground-truth-label objects as ``unknown'' at test time. Our work is a sub-problem of OWD and is focused on the \textit{localization} of all objects.

\paragraph{Localization Cues} There have been several works that have captured the notion of ``objectness'' by learning centerness \cite{tian2019fcos}, IOU prediction \cite{kim2021learning, lin2018focal, jiang2018acquisition}, or dice coefficent \cite{6909816}. These works demonstrate the strength of optimizing for object characteristics, rather than just classification loss. Centerness \cite{tian2019fcos} and IOU prediction \cite{kim2021learning, lin2018focal, jiang2018acquisition} are two particularly strong objectness measures, because their visual heatmaps have been shown to correlate to object locations. All of these works, except for Kim et. al \cite{kim2021learning}, combine the classification and localization paradigms by multiplying the logit prediction by the objectness score; however, we seek to use these cues just for ranking proposals during the localization phase.

\paragraph{Open World Detection} Joseph \etal \cite{joseph2021open} established the Open World Detection problem, as opposed to the classic open-set and open-world classification problems that came before it \cite{dhamijaopenset, openworldclass}. They propose a learning curriculum to iteratively learn unknown objects, by performing contrastive learning with an existing set of known classes; as these unknowns are classified, they are added to the known database and the process continues. This work differs from long-tail distribution works \cite{zhong2021improving, ren2020balanced, zhang2021distribution, liu2019largescale}, because while those works add an unknown label a priori, Joseph \etal \cite{joseph2021open} self-labels unknown objects during training. This work considers Open World Detection as both a localization and classification problem, but the classification task includes self-labeling these unknown objects. Conversely, Kim \etal~\cite{kim2021learning} proposed the Object Localization Network (OLN) to solve Open-World Class Agnostic Proposal Generation, which we shorten to Open-World Proposals (OWP). This is a subproblem of OWD, focused on just the proposal generation phase, and OLN used objectness cues to score proposals in a class-agnostic fashion. They were able to show that learning objectness cues was enough to generalize performance to unseen classes; however, when integrating the classification task on the seen classes, performance dropped. Since they used a two-stage architecture, simulateneously learning specific classification on the seen classes while generalizing localization to the unseen classes proved impossible, as performance dropped -6.2\%. In this work, we show that a one-stage architecture like FCOS can retain both classification and OWP performance.
\section{Our Approach}
\label{sec:our_approach}
\begin{figure}[t]
\begin{center}
    \includegraphics[width=0.75\linewidth]{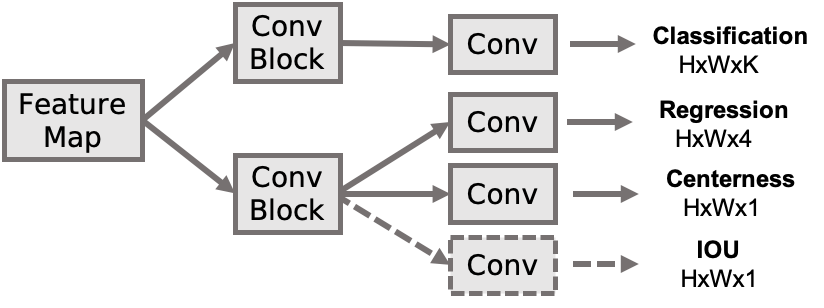}
\end{center}
\vspace{-1em}
\caption{
    Modified FCOS detection head with IOU Overlap prediction. The head takes feature maps from backbone to produce Classification, Regression, and Centerness outputs. We add the IOU branch. Conv Block consists of Convolutional layer, Group Norm, and ReLU.
}
\label{fig:arch}
\end{figure}

\paragraph{Problem Formulation} A typical object detector consists of localization task $\mathcal{L}$ and classification tasks $\mathcal{C}$, and only detect objects from the seen training class sets $C^B$. We aim at adapting these two branches and their loss functions to add \textit{open-world proposal} capability to such category-based detectors to ``localize" objects beyond the categories in the training set, such as a novel evaluation class set $C^N$. The goal is to achieve high Average Recall (AR) on $C^N$, while maintaining the category-based Average Precision (AP) on $C^B$ before adaptation.  

\subsection{Background: FCOS} 
\label{sec:fcos_background}

FCOS is a fully convolutional one-stage detection network that consists of a ResNet~\cite{he2015deep} Feature Pyramid Network (FPN) backbone \cite{lin2017feature}, and a detection head with localization, classification, and objectness branches. We denote the feature extractor as $\mathcal{F}_{\theta}(I)$, where $I \in \mathbb{R}^{H\times W\times 3}$ is the input image.  As seen in Figure~\ref{fig:arch}, the FCOS detection head outputs three main dense  predictions: \textit{Centerness} $\hat{C} \in \mathbb{R}^{H\times W}$, \textit{Bounding Box Regression} $\hat{R} \in \mathbb{R}^{H\times W \times 4}$, and \textit{Classification} $\hat{K} \in \mathbb{R}^{H\times W \times |C|}$. For a pixel located at $(x,y)$, the regression target (ground truth) is defined as $[l^*,r^*,t^*,b^*]$. Here $[l,r,t,b]$ represent the pixel distances from $(x,y)$ to the left, right, top, and bottom sides of the bounding box at that pixel. The centerness at pixel $(x, y)$'s target is computed as: 
\begin{equation}
{C^*}_{x, y} =  \sqrt{\frac{\min(l^*,r^*)}{\max(l^*,r^*)} \times \frac{\min(t^*,b^*)}{\max(t^*,b^*)}}
\end{equation}
\begin{figure}[t]
\hspace{3em}
\begin{center}
\includegraphics[width=0.85\linewidth]{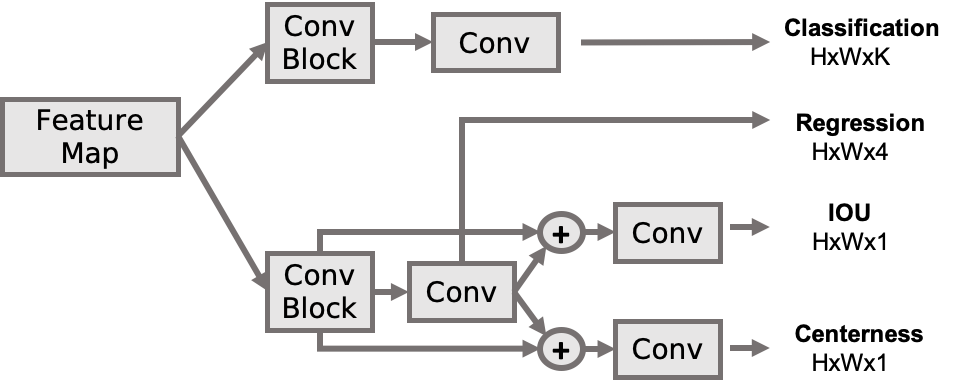}

\end{center}
\caption{
FCOS detection head with Conditional IOU and Centerness. We take the feature maps after Conv Block, concatenate with the regression branch's output and use it as input to the IOU and centerness branches.
}
\label{fig:cond_arch}
\end{figure}

\paragraph{Training} The centerness prediction branch is updated using binary cross entropy (BCE) loss, the regression branch is updated using IOU Loss \cite{yuunitbox2016}, and the classification loss is updated using Focal Loss. \cite{lin2018focal}. The centerness and regression branches are only updated using pixel locations within the ground-truth bounding boxes, while the classification loss is computed across all pixel locations in the image. To train the regression branch, \textit{centerness sampling} is applied, which samples pixels within a radius $r$ from the centers of each ground-truth bounding box in an image as positive samples~\cite{huang2015densebox, redmon2018yolov3} for both classification and regression branches. The purpose is to suppress low quality proposals.

\paragraph{Inference} The top 2000 proposals whose scores are above some defined threshold are passed to Non Maximum Suppression (NMS). Post NMS, another threshold minimum is applied, of which the top 100 proposals are used for the COCO dataset and the top 300 are used for the LVIS dataset.

\subsection{Open World Proposals (OWP)}
\label{sec:owp}

The original FCOS implementation has centerness as an objectness measure to weight the classification score as $p_{x, y} = \hat{K}_{x, y} * \hat{C}_{x, y}$. Effectively, this down-weights the prediction scores of proposals further away from the center of the ground-truth~\cite{tian2019fcos}. As a result, the low quality bounding boxes can be filtered out by a standard NMS process. To enable class-agnostic proposal generation, we can directly score proposals from the centerness prediction. This process is similar to how FCOS makes proposals from classification score, but differs in that there is only one ``class'': object.  When the centerness score is larger than a user-defined threshold, it is predicted to be an object; otherwise, it is called background. 
 
 \paragraph{IOU Overlap as Objectness}
 In practice, we observe that using centerness as an objectness cue doesn't perform well, as shown in Section~\ref{sec:experimentation}. Thus, we propose IOU score~\cite{lin2018focal, kim2021learning} between the predicted and ground-truth bounding boxes as a better objectness measure. $\text{IOU}$ is defined as the ratio of the area of intersection divided by the union of the predicted and target bounding boxes. The IOU score head, represented as $\hat{I}$, outputs a dense prediction map of size $H\times W$, whose pixel value at any pixel coordinate is defined as:
\begin{equation}
    \hat{I}_{x,y} = \text{IOU}([l,r,t,b], [l^*,r^*,t^*,b^*])
\end{equation}
The IOU function takes two regression tuples and computes their IOU as defined below:
\begin{gather*}
    I = (\min(l, l^*) + \min(r, r^*))*(\min(b, b^*) + \min(t, t^*)) \\
    U = (l^* + r^*)*(t^* + b^*) + (l + r)*(t + b) - I \\
    \text{IOU} = I/U
    \end{gather*}

We argue this is a better objectness measure since: (1) when the centerness $\hat{C}_{x, y}$ is high and the error between the ground-truth is small, if the $\hat{R}$ is off, the prediction is not as good as it is directly decided by the quality of $\hat{R}$ (2) We found it is harder to learn centerness than learning the IOU score during training, evidenced in Section~\ref{sec:owl_baseline}.
\begin{figure}[t]
\hspace{3em}
\begin{center}
\includegraphics[width=0.75\linewidth]{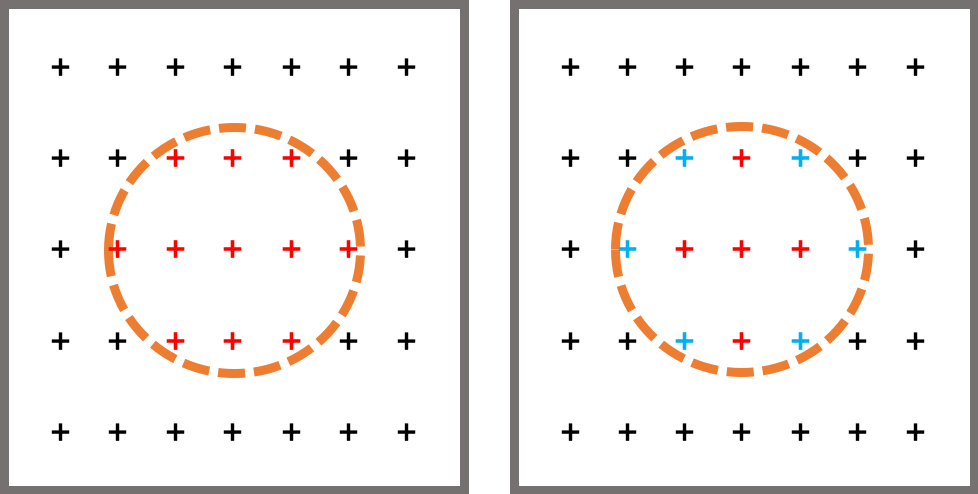}

\end{center}
\caption{
We have two ground-truths boxes, where a black + indicates a foreground sample. The orange circle represents the center-sampling, where the red points are center-sampled. On the right figure, we apply our IOU Sampling technique to create blue +, which are hard-negative samples.
}
\label{fig:vis_compare}
\end{figure}

\paragraph{Conditional IOU Head}
The aforementioned implementations make the IOU and Centerness predictions independent of the regression branch. However, we notice that the IOU and Centerness are inherently dependent on the regression branch's ground-truth target $[l^*,r^*,t^*,b^*]$ and the predicted localization $[l,r,t,b]$ at every given pixel. To provide more information to the objectness branches and improve accuracy, we investigate adding a convolutional layer that first concatenates a feature map from the Feature Pyramid $(H \times W \times 256)$ and a copy of the regression prediction $(H \times W \times 4)$ and then performs a convolution to predict the IOU/Centerness prediction per pixel $(H \times W \times 1)$, which is shown in Figure~\ref{fig:cond_arch}.

\paragraph{IOU Sampling} $\hat{I}_{x,y}$ or $\hat{C}_{x,y}$ as objectness isn't a binary classification problem; these predictions range within $[0, 1]$. Conceptually the objectness branches must see both high and low quality boxes to be able to discriminate between true objects and background. However, FCOS's existing sampling method first considers all foreground pixels, and sequentially applies \textit{Center Sampling} which samples a fixed radius of pixels from each ground-truth boxes' center. This leads to the number of positive samples vastly reduces. Overall, less positives in objectness to reweigh the $\hat{K}$ benefits the proposal selection, improving the overall accuracy. However, from the perspective of the convergence of the objectness branch, it is not beneficial to have an imbalanced ratio between positives and negatives as evidenced in Section~\ref{sec:owl_baseline}. Additionally, if we only sample points close to the center of the box, these points will always have high centerness, and thus at inference, the centerness hasn't been trained to down-weight boxes on the outskirts of the ground-truth boxes.

We need to either supply more training samples or further create granularity between the area masked and center sampled points to provide more diversity for the centerness and IOU branches. OLN \cite{kim2021learning} trains its localization branch by sampling 256 region proposals which have $\text{IOU} > r, r=0.3$. In the same vein, we propose \textit{IOU Sampling}, which sets the IOU of pixels within the ground-truth whom have IOU $<$ 0.3 to 0.0. This means that low-quality boxes or those that have little IOU with the ground-truth, will automatically be set to zero while any pixel with IOU $>$ 0.3 will retain their IOU Score. The purpose of this procedure is to flatten the IOU target distribution, which will naturally be left skewed around high IOU when the regression predictions are accurate. Our final sample procedure consists of sampling all pixels post center sampling and IOU Sampling. As shown in Figure~\ref{fig:vis_compare}, our thresholding procedure creates a more diverse training set for objectness, which we empirically show can improve OWP performance.

\begin{table*}[t]
    \centering
    \caption{
        Here we combine the results of three experiments: Baseline, OWP, OWP-Aware Classification. The Baseline experiment uses the default scoring method from FCOS, which is $\text{logits}*\text{centerness}$ and evaluates classification AP and proposal AR100 on the base class. OWP are class-agnostic proposal experiments where we seek to only maximize AR100. In OWP-Aware Classification, we seek to maximize AR100 while also maximizing classification AP. For OWP and OWP-Aware Classification, we provide the results for these experiments from OLN \cite{kim2021learning}.
    } 
    \resizebox{\linewidth}{!}{
        \begin{tabular}{ccccccc}
            \toprule
            {} & {}&  \multicolumn{3}{c}{$\text{COCO}^\text{base}$} & \multicolumn{2}{c}{$\text{COCO}^\text{novel}$}\\
            {} & {Method} &  AP &  AR10 & AR100 & AR10 & AR100 \\
            \midrule
            Baseline & $\text{logits}*\text{centerness}$\cite{tian2019fcos} & 44.21 & 55.33 & 59.23 & N/A & N/A\vspace{0.35em} \\ \hline
            \multirow{7}{*}{OWP} &centerness & N/A & 22.71 & 44.09 &  9.900 & 25.48\\
             &IOU & N/A & 32.67 & 53.26 & 13.43 & 28.31 \\
             &$\sqrt{\text{centerness}*\text{IOU}}$ & N/A & 26.20 & 48.66 & 11.60 & 28.10 \\ 
             &IOU-CS-IS & N/A & 32.86 & 53.59 & 13.08 & \bf 30.19  \\
             & Conditional IOU-CS-IS & N/A & 34.52 & 54.10 & 14.51 & \bf 31.26 \\
             & Class-Agnostic \textbf{OLN 1-RPN} \cite{kim2021learning}  & N/A &  N/A &  N/A & 11.70 & 27.40 \\
             &Class-Agnostic \textbf{OLN 2-RPN} \cite{kim2021learning} & N/A &  N/A &  N/A & 17.70 & 32.70 \vspace{0.35em}\\ \hline 
            \multirow{4}{*}{OWP-Aware Classification} &$\text{logits}*\text{IOU-CS-IS}$ (Finetune) & 46.07 & 59.25 & 64.12 & 11.67 & \bf28.39 \\
             &$\text{logits}*\text{IOU-CS-IS}$ (joint) & 46.01 & 58.24 & 63.45 & 11.12 & 26.89 \\
             &$\text{joint} + \text{unknown}$ &\bf 46.21 & \bf 60.42 & \bf 64.85 & 11.32 & 27.82 \\
             & \textbf{OLN 2-RPN} \cite{kim2021learning} & N/A & N/A & N/A & 13.30 & 26.50 \vspace{0.35em}\\ \hline
            \bottomrule
        \end{tabular}
    }
    \label{tab:all_comparison}
\end{table*}

\begin{figure}[t]
\hspace{3em}
\begin{center}
\includegraphics[width=\linewidth]{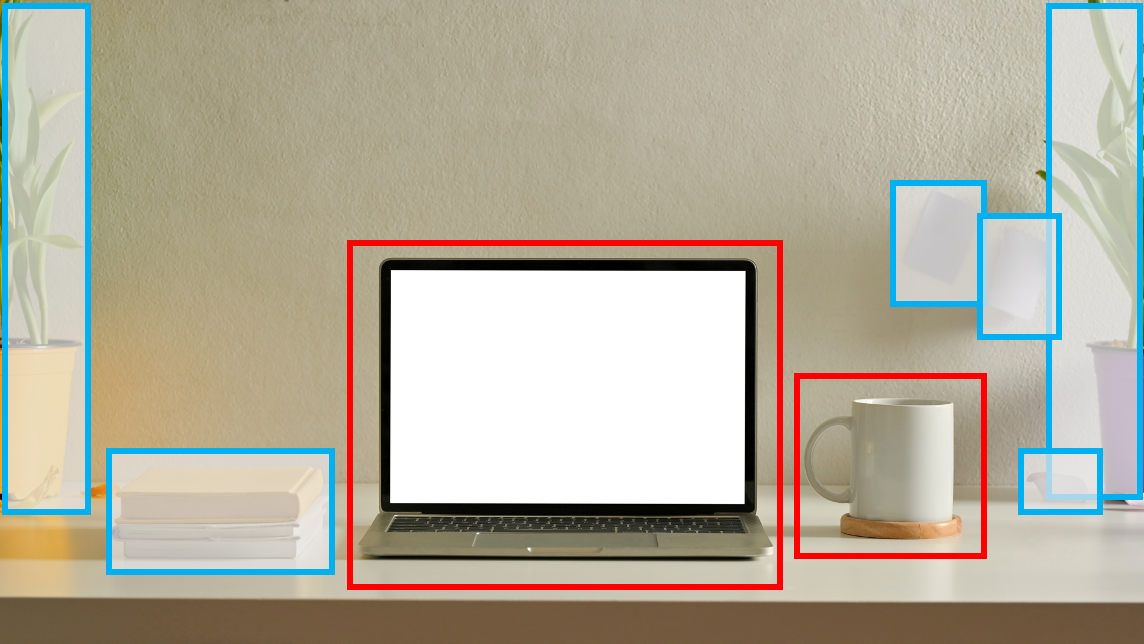}

\end{center}
\caption{
Here the red bounding boxes are the ground-truth annotations; however there are other objects on the table and on the wall that are unannotated. Therefore, we use our objectness branch to mask out the portions of those unknown objects that don't intersect with the ground-truth,  represented in blue.
}
\label{fig:unknownobj_vis}
\end{figure}
\subsection{OWP-aware Classification}
\label{sec:owp_classification}

\paragraph{Unknown Object Masking} Traditionally, detection suffers from non-exhaustive annotation \cite{gupta2019lvis, yang2020object}, which means that images aren't fully annotated for every object in view. For example, with a group of people a few of the people might be annotated while the rest are left to be considered background. Since the classification layer uses all pixels, it will be trained to classify a few of these un-annotated objects as background even though they are foreground. To alleviate this issue, we seek to use our measure of ``objectness'' from the IOU and Centerness branches, to determine ``unknown objects''. If we can properly identify the regions outside of the ground-truth that are unknown objects, we can filter out these regions before calculating the loss on the classification task, so as to minimize the number of misrepresented background samples. In a sense, the objectness branch will act like a binary mask with which to select background pixels, an example of which is shown in Figure~\ref{fig:unknownobj_vis}. We simplified the implementation of unknown object masking by masking out individual pixels whose predicted objectness was above a user-defined threshold. We tested unknown object masking with bounding boxes, more detail of which can be found in Appendix~\ref{sec:unknown_area}. 

\section{Training and Inference Recipe}
\label{sec:training}
\paragraph{OWP}
For achieving OWP on FCOS, we freeze the classification branch and only train the regression and objectness branches. Follow \cite{tian2019fcos} with centerness, the objectness branch is backpropograted with the BCE loss. Upon inference, the proposals from the regression branch are scored using the objectness branch. 

\paragraph{OWP-aware Classification}
For this, we unfreeze the classification branch and train the objectness, classification, and regression branches. During inference, proposals are scored using the multiplication of the logit predictions from the classification branch and the objectness branch predictions, as $p_{x,y}=\hat{K}*\hat{I}$. We specifically show two different variants:
\begin{enumerate}
    \item Joint-training: Joint training of the classification and regression and objectness branches. Here, each branch is given a learning rate of 0.01.
    \item Finetuning: In this setup, we pretrain the network on the pure OWP task, including the backbone and the localization branches. In the finetuning stage, the objectness branches and regression branches are given a slowed learning rate of 0.001, while the classification branch is given a learning rate of 0.01. 
\end{enumerate}

\section{Experimentation}
\label{sec:experimentation}

\paragraph{Datasets} COCO \cite{lin2015microsoft} is an object detection dataset consisting of of 330,000 images, which have annotations from 80 total classes. We split the classes into two sets: 20 seen base classes that appear in PASCAL VOC and 60 unseen novel classes that don't appear in PASCAL VOC. During evaluation, we remove all annotations belonging to the seen classes, so only unseen classes contribute to evaluation. Additionally, we evaluate on LVIS \cite{gupta2019lvis}, which is used for evaluating long-tail distributional detection. LVIS consists of 1203 classes that are split into 337 rare (r), 461 common (c), and 405 frequent (f) categories dependent on their occurrence in the dataset. The seen class (base) consists of all rare classes, common classes, and the first 305 frequent classes, while the unseen class (novel) consists of the rest of the 100 frequent classes. This makes the seen (base) and unseen (novel) categories disjoint.

\paragraph{Evaluation Metrics} For OWP, we measure the Average Recall (AR\underline{$N$}) over $N$ proposals on the novel classes. For COCO $N=100$ and for LVIS $N=300$. For the OWP-Aware Classification task, we measure Average Precision (AP), which is averaged over each category in the novel set.

\paragraph{Implementation Details} All experiments are trained at a batch size of 16 across 8 GPU's for 90,000 epochs, using stochastic gradient descent with momentum of 0.9 and weight decay of 0.0001. The default learning rate across all experiments is 0.01, unless further specified in the OWL-aware classification experiments. This learning rate is decayed by a multiplicative factor of 0.1 at 60k and 80k epochs.

\subsection{OWP Benchmarking}
\label{sec:coco_benchmarking}

\paragraph{Baselines}\label{sec:owl_baseline} We first perform a standard object detection training--joint training of the classification and localization branches and we evaluate per-category $AP$ and $mAP$, which uses $\hat{K}*\hat{C}$ for proposal scoring on the base classes. This is shown in the first row of Table~\ref{tab:all_comparison}. We perform this experiment to compare AP and AR100 statistics on the base classes.

\paragraph{OWP} We train three permutations of class-agnostic FCOS with different measures of objectness -- centerness, IOU, and $\sqrt{\text{centerness}*\text{IOU}}$. \textbf{The goal in OWP is achieve the highest possible AR100 on both the novel and the base classes}. As seen in Table \ref{tab:all_comparison}, in the OWP section, using centerness, which is the FCOS's\cite{tian2019fcos} primary scoring mechanism for scoring, has \textbf{-3.44\%} AR100 scores compared to IOU on the novel classes, which indicates IOU is a stronger objectiveness objective than centerness.
\begin{figure}[t]
    \centering
    \subfigure[
        $\text{logits}*\text{centernesss}$ proposal scoring distribution
    ]{
        \includegraphics[width=0.9\linewidth]{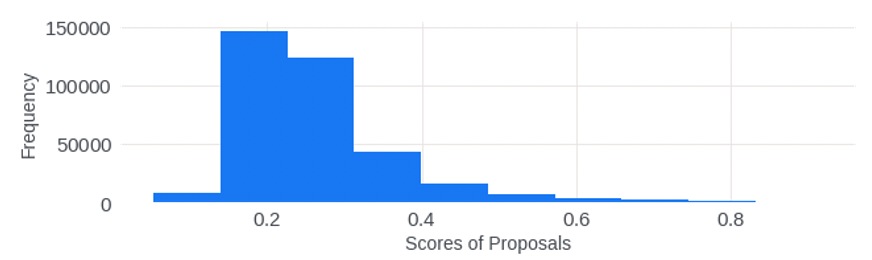}
        \label{fig:fcos_prob1}
    }
    \subfigure[
        Centerness proposal scoring distribution
    ]{
        \includegraphics[width=0.9\linewidth]{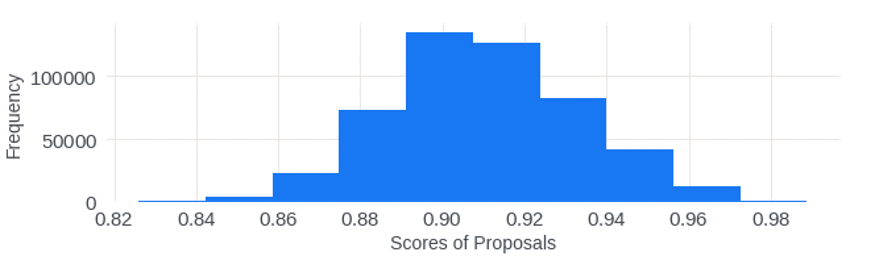}
        \label{fig:fcos_prob2}
    }
    \subfigure[
        IOU proposal scoring distribution
    ]{
        \includegraphics[width=0.9\linewidth]{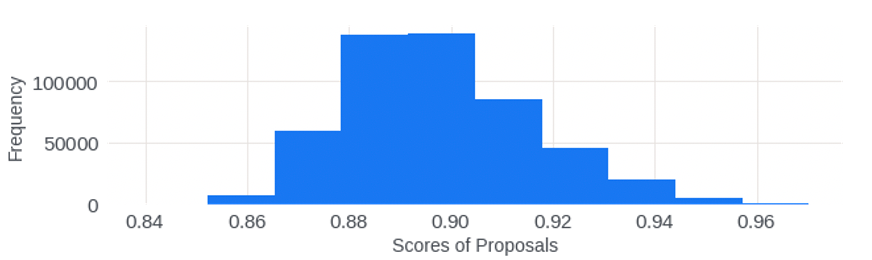}
        \label{fig:fcos_prob3}
    }
    \caption{Scoring Distribution for Objectness}
    \label{fig:all_probs}
\end{figure}
Additionally, $\text{logits}*\text{centerness}$ as expected outperforms the other pure objectness-based measures by a significant margin on the base classes. We further investigated this phenomena by first analyzing the histogram scoring distributions for $\text{logits}*\text{centerness}$, centerness, and IOU, in Figure~\ref{fig:all_probs}. While not easily noticeable, the distribution in Figure ~\ref{fig:fcos_prob3} is slightly more skewed to the left than Figure~\ref{fig:fcos_prob2}. The most left-skewed scoring is $\text{logits}*\text{centerness}$, as seen in Figure~\ref{fig:all_probs}. It seems that the benefit with using logits is that high scoring proposals occur with much lower frequency than low-scoring proposals. This leads to a higher \textit{granularity} between non-objects and true-objects, which neither centerness nor IOU seem to fully capture. Diving deeper, we find the loss convergence of centerness in this experiment is not meaningful, as it starts at 0.636 and decreases to 0.612; however, for IOU, the loss converges from 0.62 to 0.5. Therefore, IOU seems to perform better as a objectness metric, because it converges during training and has a more left-skewed distribution than centerness. However, since IOU still doesn't match state-of-the-art, we apply our sampling optimizations.

In the method labeled IOU-CS-IS in OWP of Table~\ref{tab:all_comparison}, we find that the combination of Center Sampling (CS) and IOU Sampling (IS) for the IOU branch achieves superior performance to centerness and $\sqrt{\text{centerness}*\text{IOU}}$. Additionally, we provide the results of \textbf{OLN 1-RPN} \cite{kim2021learning} and \textbf{OLN 2-RPN} \cite{kim2021learning}. \textbf{OLN 1-RPN} consists of the proposal stage of the OLN architecture with centerness as a box quality measure, while \textbf{OLN 2-RPN} consists of both stages of the OLN architecture with $\sqrt{\text{centerness}*\text{IOU}}$ as an objectness measure. We find that our IOU Sampling method with IOU has better AR100 performance than \textbf{OLN 1-RPN} and only 2\% lower AR100 performance than \textbf{OLN 2-RPN}. This signifies that objectness is working and effective for OWP.

We get even closer to OLN 2-RPN with our Conditional IOU Architecture, where we condition the Centerness and IOU branches on the regression prediction and feature pyramid. In the method labeled Conditional IOU-CS-IS of Table~\ref{tab:all_comparison}, this architecture is the best performer, achieving \textbf{31.26\%} AR100.

\subsection{OWP-Aware Classification Benchmarking}

\paragraph{OWP-Aware Classification}
\label{sec:owlawareclass}
From here, we seek to understand how adding the classification task back in might affect OWP performance. Kim et. al \cite{kim2021learning} found that adding classification into the second stage of the OLN architecture caused AR100 performance drop from \textbf{32.7} to \textbf{26.5} (6.2\%), as seen in OWP-Aware Classification for OLN 2-RPN. Therefore, with our OWP optimizations, we show two variants to retain classification performance in FCOS, joint, and finetune. The full results can be found in Appendix~\ref{sec:training_recipe}. \textbf{The goal of OWP-aware Classification is to retain AR100 on the novel class, while still achieving a high AP on the base class}. We provide the results of finetune and joint training in the OWP-Aware Classification section of Table~\ref{tab:all_comparison}. We find that finetuning is the best at retaining OWP recall, only dropping 2\% from the IOU-CS-IS, and matches joint training AP at 46\%. Joint training is more preferred, as it is more time-efficient and simpler to engineer, so therefore, we investigate unknown object masking.

Unknown object masking relies upon making a binary decision of whether a background pixel belongs to an unknown object or is truly background. This requires a threshold to be defined before-hand which relays the minimum confidence needed for a pixel to be considered foreground. As previously seen in Figures \ref{fig:fcos_prob1}, \ref{fig:fcos_prob2}, \ref{fig:fcos_prob3}, we saw that proposals were normally distributed around an IOU value of 0.90, so we decided to test two thresholds: 0.925 and 0.95. However, we also tested when to apply these thresholds, because unknown object masking also relies upon the box quality estimates to be accurate. We test the AP on COCO Base Classes and AR100 on COCO Novel Classes at different thresholds: [0.925, 0.95] and different epochs: [5k, 10k, 30k, 60k]. We found that a threshold of 0.95, applied at 5k epochs can benefit joint training (our full results are in Appendix~\ref{sec:training_recipe}). This is also more desirable, since this setup also achieves the highest AP on the classification task. Therefore, as seen in the OWP-Aware Classification section for method ``joint+unknown'' of Table~\ref{tab:all_comparison}, we can achieve an improvement in AP from \textbf{46} to \textbf{46.2\%}, while increasing AR100 from \textbf{26.89} to \textbf{27.8\%}. This is most indicative of our architecture's benefits relative to OLN 2-RPN\cite{kim2021learning}, because OLN 2-RPN drops \textbf{6.2\%} in the OWP-Aware Classification Task, shown in the last row of the OWP-Aware section of Table~\ref{tab:all_comparison}, whereas our architecture only drops \textbf{2.3\%} while achieving high AP. The decoupling of the localization and classification outputs allow for both OWP and classification to be optimized, which a two-stage architecture like OLN 2-RPN provably struggles with.

\subsection{LVIS Benchmarking}

\paragraph{LVIS}
\label{sec:lvis_benchmarking}
Additionally, we sought to understand how the OWL and OWL-aware classification tasks might transfer to the LVIS\cite{gupta2019lvis} Dataset. We train on the base classes and evaluate on the novel class for four different settings; all settings are the same as on COCO dataset except that on LVIS the detection output 300 detection results instead of 100 as on COCO dataset. We chose to analyze the baseline FCOS performance on LVIS, IOU-CS-IS, joint, and joint+unknown training, which are the best performing optimizations seen in COCO.
\begin{table}[t]
    \centering
    \caption{
        Evaluating the Impact of IOU, IOU Sampling, and Unknown Object Masking on Classification and OWD in LVIS.
    } 
    \resizebox{\linewidth}{!}{ 
        \begin{tabular}{ccccc}
            \toprule
            {} & {Method} & \multicolumn{2}{c}{$\text{LVIS}^\text{base}$} & \multicolumn{1}{c}{$\text{LVIS}^\text{novel}$} \\
            {} & {} & AP & AR@300 & AR@300 \\
            \midrule
            \multirow{1}{*}{Baseline} & $\text{logits}*\text{centerness}$ & 18.88 & 27.77  & 35.30\\ \hline
            \multirow{1}{*}{OWP} & IOU-CS-IS & N/A & N/A & 39.85 \\ \hline
            \multirow{2}{*}{OWP Classification} & joint & 18.84 & 27.29 & 39.28 \\ 
             & joint+unknown & 18.56 & 27.25 & 39.21 \\

            \bottomrule
        \end{tabular}
    }
    \label{tab:lvis}
\end{table}

An encouraging sign from the LVIS testing is that our OWL setup with Center Sampling and IOU Sampling increases AR300 from 35.30\% to 39.85\%, which indicates that our box quality metric is truly working, because we saw similar gains in COCO. Additionally, during OWL-aware Classification, the AP doesn't significantly drop from the original FCOS AP (18.88 to 18.84\%); however, we still maintain the OWL gains from class-agnostic training. Unfortunately, we don't see any significant gains from unknown object masking in the LVIS dataset, which was unexpected, since LVIS has a large number of unknown objects in the background. Further work needs to be done to understand why the unknown object masking procedure doesn't translate to LVIS, and this is discussed in Appendix~\ref{sec:unknown_area}.

\section{Conclusions}
We provide a thorough investigation of the FCOS Architecture on OWP and OWP-aware classification. Similar to OLN \cite{kim2021learning}, we investigate the baseline performance of FCOS on the COCO Dataset and find that its default centerness mechanism isn't very effective for OWP. Therefore, we investigate using IOU as an objectness measure with an IOU Sampling mechanism to enforce hard negatives, and these optimizations improve OWP by \textbf{6}\%. Additionally, we added the classification task back in and discovered that due to the decoupling of the classification and regression heads, FCOS can retain OWP performance, while still maintaining classification accuracy, better than OLN. We also propose Unknown Object Masking to remove ``unknown objects'', which improved performance in both OWP and classification. For robustness, we carry all our optimizations to the LVIS dataset, where we see performance enhancements for every optimization, except for unknown object masking. Further research will consist of improving the unknown object masking procedure by using non-maximum suppressed areas, rather than individual points. In conclusion, we show that FCOS is successful in retaining classification performance while achieving OWP, which is the first one-stage network to do so.

\subsection*{Acknowledgements}
We would like to thank Weiyao Wang and Haoqi Fan for their help with the architecture design and class-agnostic evaluation code. We also thank Zhicheng Yan for overseeing the project and monitoring performance.

{
    \small
    \bibliographystyle{ieee_fullname}
    \bibliography{macros,main}
}

\appendix

\setcounter{page}{1}

\twocolumn[
\centering
\Large
\textbf{Appendix} \\
\vspace{0.5em}Supplementary Material \\
\vspace{1.0em}
] 
\appendix
\begin{figure*}[t]
    \centering
    \subfigure[
        AP over Epochs for COCO Base @ Threshold=0.925
    ]{
        \includegraphics[width=0.48\linewidth]{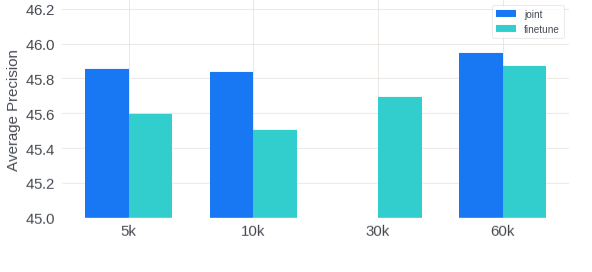}
        \label{fig:unknown_ap_925}
    }
    \subfigure[
        AR100 over Epochs for COCO Novel @ Threshold=0.925
    ]{
        \includegraphics[width=0.48\linewidth]{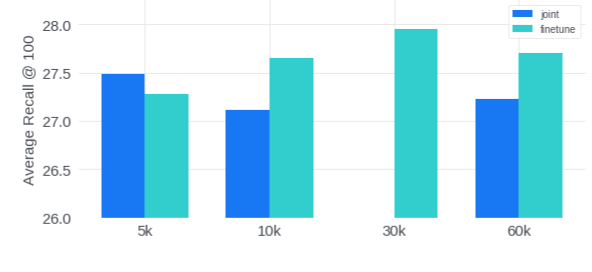}
        \label{fig:unknown_ar_925}
    }
    \subfigure[
        AP over Epochs for COCO Base @ Threshold=0.95
    ]{
        \includegraphics[width=0.48\linewidth]{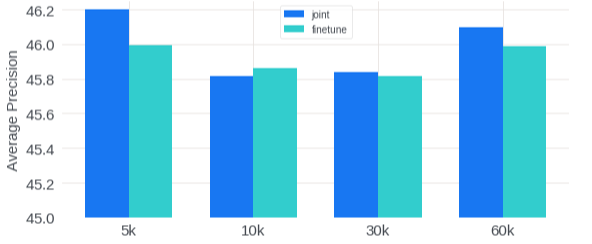}
        \label{fig:unknown_ap_95}
    }
    \subfigure[
        AR100 over Epochs for COCO Novel @ Threshold=0.95
    ]{
        \includegraphics[width=0.48\linewidth]{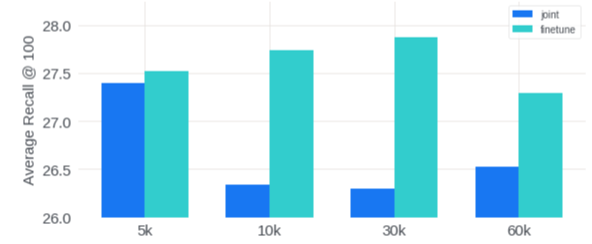}
        \label{fig:unknown_ar_95}
    }
    \caption{AP/AR100 Comparison between Various Epochs and Thresholds}
    \label{fig:unknown_obj}
\end{figure*}

\begin{figure*}[t]
    \centering
    \subfigure[FCOS Sampling]{
        \includegraphics[width=0.48\linewidth]{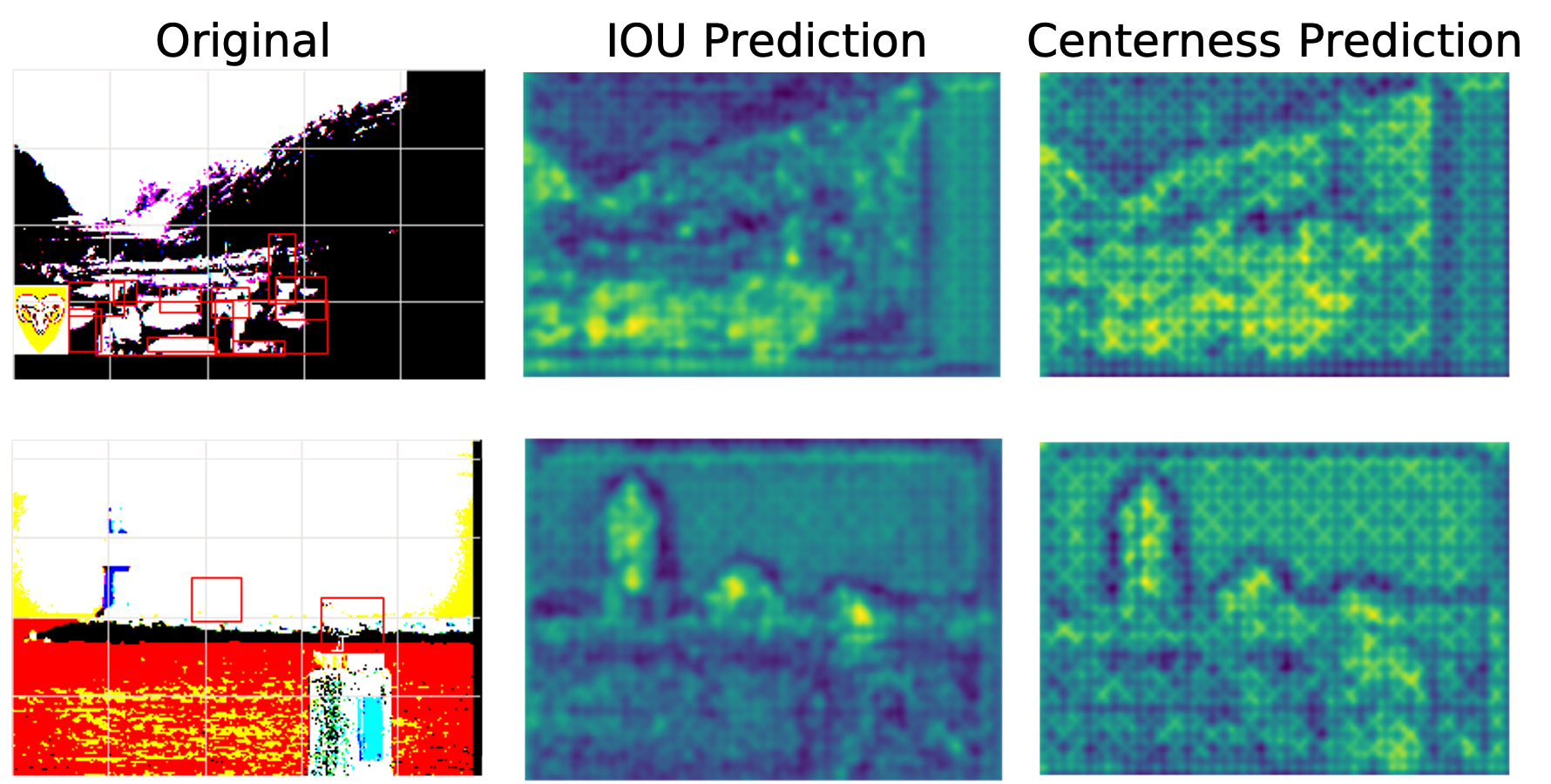}
        \label{fig:fcos_sampling}
    }
    \subfigure[All Sampling]{
        \includegraphics[width=0.48\linewidth]{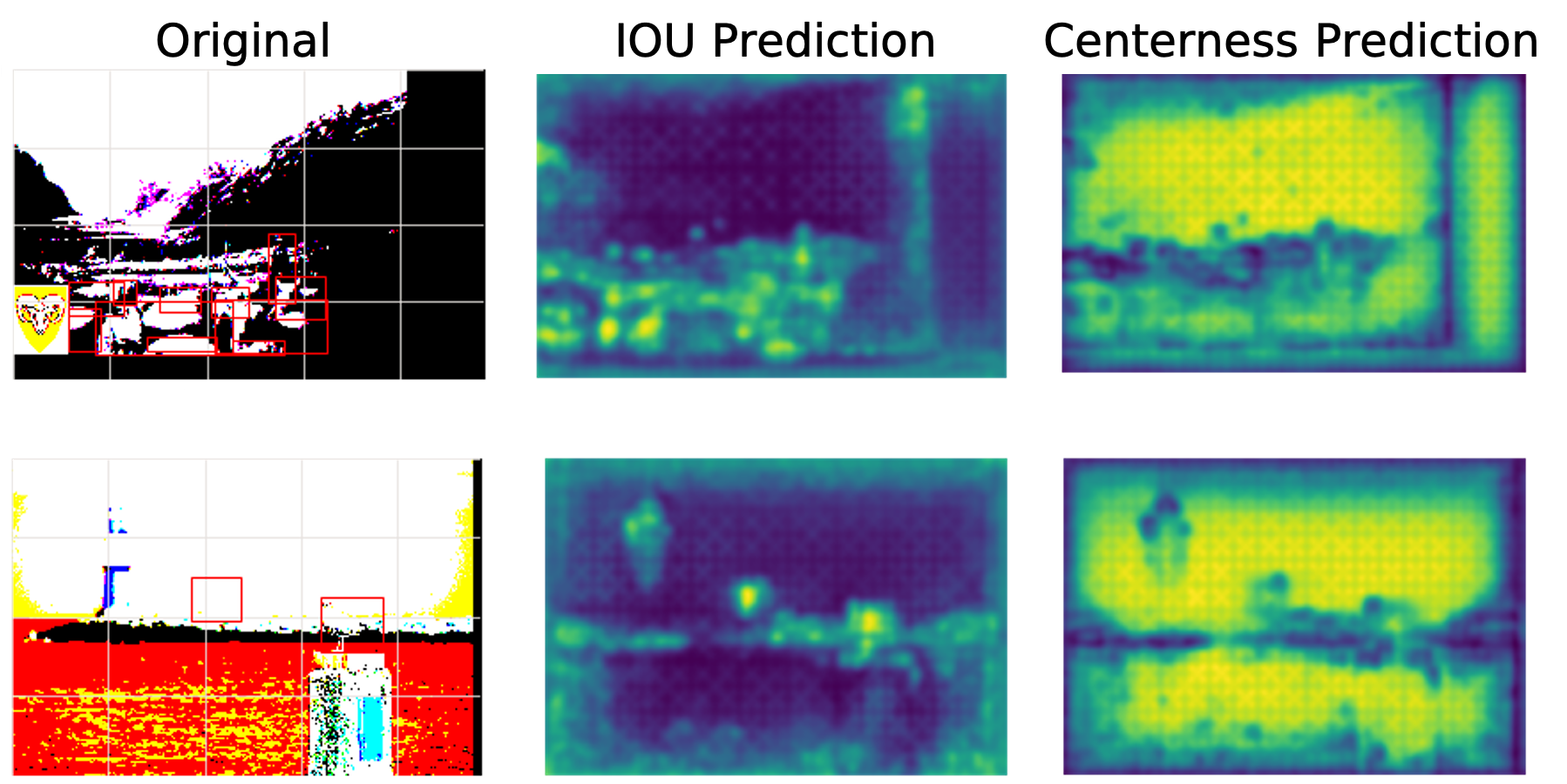}
        \label{fig:all_sampling}
    }
    \caption{Centerness and IOU Maps from FCOS Sampling vs All Sampling}
    \label{fig:compare_sampling}
\end{figure*}

\section{OWP-Aware Classification Extensions}
\paragraph{Training Recipe Testing} \label{sec:training_recipe}Earlier, we presented the results for joint and finetune training recipes; however, in this section we provide results for joint, finetune, and frozen training recipes, while also investigating whether a frozen backbone will help performance (indicated by the fb suffix). The frozen recipe is similar to finetuning, but instead of a reduced learning rate on the localization branches, we freeze the localization branches completely. In Table~\ref{tab:classification}, we see that using a frozen backbone allows (finetune frozen fb) to retain OWL performance (30.19\%), but the AP is lacking. With an unfrozen backbone, we are able to achieve ~46\% AP with joint-training and (finetune slowed), which have AR100 scores of 26.89\% and 28.39\% respectively. Notably, with our (finetune slowed) training scheme with IT and CS we are able to achieve better OWL to classification performance than OLN (26.5\%). However, we should also note that joint training is only 1.5\% off from (finetune slowed), which is an encouraging sign that adding the classification task in one-stage-detection isn't detrimental to OWL performance. Intuitively, this makes sense because in one-stage-detection networks, like FCOS, the classification and regression/box quality branches are decoupled, so gradients from on one branch don't impact the other branches like in two-stage-detection networks.

\paragraph{Unknown Object Masking Testing}\label{sec:all_unknown_obj} In unknown object masking we test thresholds [0.925, 0.95], applied at epochs [5k, 10k, 30k, 60k], and for finetune and joint training models. The highest AR100 performance is with fine-tuning and 0.925 masking occuring at 30k epochs, as seen in Figure~\ref{fig:unknown_ar_925} . However, we also notice that joint training can achieve similar AR100 performance to fine-tuning at 5k epochs with a 0.95 unknown object masking threshold. This is also more desirable, since this setup also achieves the highest AP on the classification task.

\begin{table}
    \centering
    \resizebox{\linewidth}{!}{ 
        \begin{tabular}{lrrrrrr}
            \toprule
            {} & \multicolumn{3}{c}{$\text{COCO}^\text{base}$} & \multicolumn{3}{c}{$\text{COCO}^\text{novel}$} \\
            {} &                      AP &  AR10 & AR100 &                         AP &  AR10 & AR100 \\
            \midrule
            joint &                   \textbf{46.01} & 58.24 & 63.45 &                       1.43 & 11.12 & \textbf{26.89} \\
            frozen &                   38.24 & 50.73 & 55.35 &                       0.71 &  8.27 & 20.74 \\
            finetune &                   \textbf{46.07} & 59.25 & 64.12 &                       1.59 & 11.67 & \textbf{28.39} \\
            joint fb &                   25.31 & 42.11 & 47.37 &                       0.76 &  6.45 & 19.31 \\
            frozen fb &                   38.77 & 56.54 & 61.41 &                       1.94 & 13.08 & \textbf{30.19} \\
            finetune fb &                   38.96 & 57.00 & 61.86 &                       1.94 & 12.98 & 29.68 \\
            \bottomrule
        \end{tabular}
    }
    \caption{
        Comparison of joint, frozen, and finetune variants on classification.
    } 
    \label{tab:classification}
\end{table}
\label{sec:unknown_area}
\paragraph{Unknown Area Masking} Unknown object masking was supposed to mask out background pixels that had object-like traits as to improve classification performance. However, as we saw in the previous section, unknown object masking didn't have the expected performance gains in LVIS. Perhaps only masking pixels according to the unknown object masking threshold is too sparse, and contiguous areas are necessary. Or, since the masking procedure is dependent on the accuracy of the box quality scores, this might mean that our box quality scores aren't working as well as previously expected.

We will first investigate using areas, rather than individual pixels as unknown objects. First, we find pixels that are outside of the ground-truths that have high box quality. Then we use the regression prediction at that particular pixel, which is an [l,r,t,b] tuple, to mask out any pixels within the suggested box. This would contiguously mask out pixels that exist within unknown objects. The only issue with this approach is that since high-value pixels should theoretically exist within close vicinity of one another, there will be multiple boxes that are masked within the same vicinity. This could be alleviated with Non-Maximum Suppression; however, for simplicity, we seek to understand how the basic version of unknown area masking might add performance enhancements.

\begin{table}
    \centering
    \resizebox{\linewidth}{!}{ 
        \begin{tabular}{llrrrr}
            \toprule
            {} & \multicolumn{2}{c}{$\text{LVIS}^\text{base}$} & \multicolumn{2}{c}{$\text{LVIS}^\text{base}$} \\
            {} &                       AP & AR@300 &                            AP & AR@300 \\
            \midrule
            areas\_0k &                    16.58 &  24.88 &                          2.24 &  38.74 \\
            areas\_5k &                    16.36 &  24.67 &                          2.28 &  38.96 \\
            areas\_10k &                    16.41 &  24.72 &                          2.31 &  39.01 \\
            areas\_60k &                    16.38 &  24.77 &                          2.29 &  38.86 \\
            \bottomrule
        \end{tabular}
    }
    \caption{
        Unknown Area Masking Evaluation
    } 
    \label{tab:unknown_area}
\end{table}

Unfortunately, this does not yield any benefits in AP or AR300 performance relative to the previous results, so like the previous section, further work needs to be conducted to evaluate the effectiveness of unknown masking.

\section{All Sampling}
We seek to investigate how using ALL pixels, rather than the area-masked, center sampled pixels may affect objectness performance. Intuitively, while it make sense that we should center sample pixels and area mask during regression, we seek for the objectness measures to be able to discern between both low and high box-quality pixels. If we only use center-sampled/area-masked pixels, only pixels towards the center will be used, which therefore means only high-magnitude pixels will be used to train the IOU and centerness branch. Binary classifiers generally need a one to one ratio of positive and negative samples, and if we consider any pixel with box quality score greater than 0.5 a positive sample, then both IOU and centerness pixel training sets are likely to only have positive samples present. To encourage a spread of positive and negative samples, we sought to understand how training on ALL pixels might improve performance. Interestingly, this change kept the AR100 relatively the same on the OWL task (coco base class agnostic AR100) with a score of \textbf{30.18}. Qualitatively though, this has a beneficial impact on the visualizations of the objectness maps.

In Figure ~\ref{fig:compare_sampling}, we see the visualized IOU and centerness maps for FCOS Sampling vs All Sampling for two images. We have two images, the first of which has a large number of small boxes near the bottom of the image, while the second image has two medium-sized bounding boxes near the center of the image. Both Figure 3a and 3b only train the IOU branch, but we visualize both the iou and centerness maps. With FCOS Sampling and All Sampling, we have high-density, high-valued pixels (represented by the yellowed colors) near the foreground images. Interestingly, we see on the second image, that both also pick up on the unknown foreground object (which is the leftmost high density, high-valued cluster). However, a considerable difference between the two Sampling methods, is that FCOS Sampling attributes high-density, high-valued pixel clusters to objects, but high-values sparse pixels are sprinkled around regions in both images. With All Sampling, we see high-density, high-valued pixels at objects, but the background is completely black. This goes to show that training with all pixels allows the box quality branch to \textit{discriminate} between object and non-object pixels. But why do the two have the same AR100 Performance? Inherently, with a statistic like AR100, more proposals can benefit performance, because there is more opportunity to intersect the ground-truth with more proposals. So while those proposals may be less accurate, just by virtue of there being more proposals presents an opportunity for FCOS Sampling to match the performance of All Sampling. Better thresholding pre-and-post NMS may present better opportunity to utilize the All Sampling Method, because it is clear that it visually works better than FCOS Sampling for training the box quality branch. An interesting aside with both sampling methods is that even though we are only training IOU, centerness shows the same heatmap visualization features as IOU. This indicates that the down-stream features (FCOS box share tower and trainable backbone) are learning objectness-features.


\end{document}